\documentclass[journal]{IEEEtran}
\usepackage{graphicx}
\graphicspath{{../figures/final/}}
\usepackage{booktabs,tabularx,array,multirow,makecell}
\usepackage{amsmath,amssymb,amsthm,mathtools}
\usepackage{url}
\usepackage[hidelinks]{hyperref}
\usepackage{xcolor}
\usepackage{microtype}
\usepackage{balance}

\definecolor{deepblue}{HTML}{173B57}
\newcommand{\mode}[1]{\ensuremath{\mathsf{#1}}}
\newcommand{\silent}{\mode{silent}}
\newcommand{\ask}{\mode{ask}}
\newcommand{\assist}{\mode{assist}}
\newcommand{\act}{\mode{act}}
\newcolumntype{Y}{>{\raggedright\arraybackslash}X}
\newtheorem{definition}{Definition}
\newtheorem{proposition}{Proposition}

\title{Proactive Service Agents: A Unified Decision Framework, Methods, and Evaluation}

\author{Yan~Tang, Tingyu~Cao, Yuanbo~Tang, Huaze~Tang, and Keer~Hu%
\thanks{The authors are with the UniCone Team (https://www.unicone.club). Corresponding author: Tingyu Cao (cao-ty25@mails.tsinghua.edu.cn).}}

\begin{document}
\maketitle

\begin{abstract}
Large language model agents can plan, invoke tools, and modify external states, yet most systems still take an explicit user instruction as a fixed starting point. Proactive service moves the decision upstream: an agent must infer service opportunities from incomplete environmental and user signals, choose among remaining silent, asking, assisting, and acting, and account for interruption, misunderstanding, overreach, and privacy costs. This survey gives an operational definition centered on initiative and formulates the problem as a partially observable sequential decision process constrained by authorization and risk. The formulation represents timing, content, and delivery within one structured action, while making explicit the option value of waiting, the decision value of questions, and feedback-induced state changes. On this basis, we organize existing methods along one decision pipeline (state and need estimation, intervention gating, action construction, and feedback adaptation) and describe prescribed, predictive, model-based, and return-optimizing mechanisms as non-exclusive policy-construction components. We further normalize decision units and three-axis evidence descriptors across streaming dialogue, screen, video, software-engineering, and human--agent collaboration resources, and formalize metrics for triggering, timing, calibration, user burden, safety, and policy value. The synthesis shows why offline classification performance alone does not predict deployment benefit and why long-term memory is not a defining condition of proactivity. Reliable proactive service instead requires calibrated incremental intervention value, verifiable authorization, recoverable execution, and counterfactual evidence.
\end{abstract}


\section{Introduction}

Tool-augmented large language models (LLMs) have advanced assistants from text generators to agents that browse, program, operate graphical interfaces, and coordinate multistep tasks. This growth in capability has not resolved a more basic question: who first decides what should be done? Common evaluations begin with a complete instruction and ask a system to plan toward a known task. In sustained use, however, a need may instead appear as a screen event, a sensor stream, a historical preference, or stalled task progress, and the user may not notice or articulate it in time. Allowing a system to initiate help can shorten the path from need to service, but can also cause false alarms, interruption, manipulation, privacy exposure, and irreversible actions. Proactive service is therefore not more frequent action. It is the decision, under uncertainty, of when the system should assume initiative, together with evidence that doing so is more valuable than waiting.

The foundations of this problem are substantial but fragmented. Mixed-initiative interaction frames transfers of control between human and system as a tradeoff among expected benefit, attentional state, and interruption cost \cite{walker1990,horvitz1999mixedinitiative,horvitz1999priorities,horvitz2003attention}. Proactive dialogue studies goal guidance, clarification, policy control, and long-horizon conversational outcomes \cite{wu2019duconv,deng2023survey,deng2025proactivecai,deng2024humancentered}. Conversational recommendation, information retrieval, and just-in-time adaptive intervention respectively study preference elicitation, whether to ask, and context-dependent triggering \cite{liu2020durecdial,jannach2021crssurvey,aliannejadi2019qulac,aliannejadi2021clariq,nahumshani2018jitai}. Recent proactive agents extend these decisions to screen logs, egocentric video, longitudinal records, and tool use \cite{lu2024proactiveagent,proagentbench2026,eyeswideopen2025,pask2026}. These literatures use different temporal granularity, silent negatives, feedback sources, and risk assumptions. Organizing them application by application obscures the object that is actually comparable: the belief state, feasible actions, and consequences at each \emph{decision opportunity}.

This survey reorganizes proactive service around that object and makes three contributions. \textbf{First}, it provides an operational definition relative to the current task specification and develops a constrained partially observable sequential decision model. Silence, inquiry, assistance, and execution belong to one action space, so timing, content, and delivery can be analyzed jointly. \textbf{Second}, it synthesizes methods along one decision pipeline (state estimation, intervention gating, action construction, and feedback adaptation) while using the orthogonal dimension of how a policy is constructed to describe prescribed, predictive, model-based, and return-optimizing components. This avoids mixing model mechanisms, training paradigms, tasks, and applications at one taxonomic level. \textbf{Third}, it proposes comparable resource fields, metrics, and a three-axis evidence descriptor separating interaction realism, comparison design, and human-outcome horizon. It explains why trigger F1, content quality, or acceptance rate alone cannot represent deployment value, and incorporates authorization, reversibility, long-term utility, and counterfactual evaluation into one protocol. The supplementary review protocol records scope, selection, evidence-use, and version rules.

\subsection{Relation to Prior Surveys}

Proactivity has already been surveyed, but almost entirely within natural-language dialogue, and clarifying what those surveys settled (and where they stop) motivates the decision-centered synthesis developed here. The comparison below contrasts the most relevant surveys with this review along scope, organizing axis, notion of proactivity, evaluation, and safety.

\textbf{Proactive dialogue as an established object.} A first line of work established proactivity as a first-class property of dialogue systems. Deng et al.\ contributed the earliest survey of proactive dialogue and organized the field by dialogue type (open-domain, task-oriented, and information-seeking), cataloguing subtasks such as topic planning, non-collaborative negotiation, clarification, and preference elicitation together with their datasets and metrics \cite{deng2023survey}. Its journal extension formalized proactivity through anticipation, initiative, and planning and consolidated methods, resources, and open problems, including proactivity in LLM-based and hybrid dialogues, evaluation, and ethics \cite{deng2025proactivecai}. These surveys turned scattered tasks into a coherent research object and remain the standard reference for conversational proactivity.

\textbf{From capability to human cost.} A second, human-centered line shifted attention from what a proactive system can do to what it should do for the user. Deng et al.\ proposed the intelligence--adaptivity--civility taxonomy and reread the literature across five construction stages, arguing that initiative without restraint is perceived as intrusive \cite{deng2024humancentered}. Adjacent to dialogue, a survey of conversational recommendation systematized preference elicitation, dialogue management, and evaluation for systems that ask in order to recommend \cite{jannach2021crssurvey}. Together these surveys introduced user burden, timing, and trust as design concerns and mapped the mechanisms closest to proactive service.

\textbf{Four limitations relative to current proactive-service agents.} Read against the systems now appearing on screens, wearables, robots, graphical interfaces, and code repositories, these surveys share four limitations. \emph{First}, their scope is conversational: the decision unit is a dialogue turn and the system is usually assumed to already hold a speaking turn, whereas proactive service must decide whether and when to move from silence to intervention in an open stream where most moments carry no service event. \emph{Second}, they organize the field by dialogue type or design dimension, which keeps model mechanism, task, and application at one level and does not isolate the object that is comparable across settings: the belief, admissible actions, and consequences at each decision opportunity, together with the option value of remaining silent. \emph{Third}, they treat proactivity as a capability (leading a conversation toward a goal) rather than as a gated choice under a genuine silence alternative, and therefore leave underspecified when not to intervene, the value of waiting and asking, and authorization as a hard constraint rather than a soft penalty. \emph{Fourth}, they name evaluation and ethics as open challenges but stop short of an operational protocol: they do not supply a common decision-unit schema, calibration, timing, and coverage--risk metrics, off-policy or counterfactual estimation, or a safety model that lives in the action space, and they largely predate the 2025--2026 wave of screen, egocentric-video, wearable, and software-engineering proactive agents. These are gaps in coverage and formalization, not defects of the surveyed work, which answered the questions salient at the time.

\textbf{What this review adds.} This review responds to these limitations by changing the object of organization rather than adding another application list. It replaces the conversational scope with a modality-agnostic decision object; application-by-application cataloguing with one decision pipeline and an orthogonal policy-construction axis; a capability definition with an instruction-relative, silence-gated definition embedded in a constrained partially observable decision model; and open-ended calls for better evaluation with concrete decision units, calibration, timing, coverage--risk and off-policy metrics, an evidence descriptor, and a safety model tied to authorization and reversibility. To our knowledge, no prior survey organizes proactive service across digital-workspace, embodied, and high-stakes human-service settings under a single set of decision variables; the remainder of the paper develops that account.

\begin{table*}[t]
{\footnotesize\textbf{Representative surveys of proactivity and adjacent areas contrasted with this review.} ``Dialogue turn'' and ``decision opportunity'' denote the comparison unit; the last two columns record whether an operational evaluation protocol and an action-space safety model are provided.\par}
\smallskip
\centering
\footnotesize
\setlength{\tabcolsep}{3pt}
\renewcommand{\arraystretch}{1.05}
\begin{tabularx}{\textwidth}{@{}p{1.85cm}p{2.35cm}p{2.5cm}p{2.95cm}Yp{2.15cm}@{}}
\toprule
Survey & Scope & Organizing axis & Notion of proactivity & Evaluation treatment & Safety and authorization \\
\midrule
Deng et al., IJCAI 2023 \cite{deng2023survey} & Text dialogue (open-domain, task-oriented, information-seeking) & Dialogue type and subtask & Leading a dialogue to a preset target or goal & Per-task datasets and metrics; ethics as a challenge & Discussed as an open challenge \\
Deng et al., ACM TOIS 2025 \cite{deng2025proactivecai} & Text dialogue, including LLM proactive dialogue & Anticipation, initiative, planning $\times$ dialogue type & Taking initiative and anticipating long-term impact in dialogue & Consolidated per-task metrics; evaluation named an open problem & Ethics as a future direction \\
Deng et al., SIGIR 2024 (position) \cite{deng2024humancentered} & Conversational agents & Intelligence, adaptivity, civility across five stages & A human-centered capability balancing goal and user & Design principles per stage; no unified protocol & Civility as a design dimension \\
Jannach et al., ACM CSUR 2021 \cite{jannach2021crssurvey} & Conversational recommendation (adjacent) & System components and interaction flow & Asking to elicit preferences and guide choice & Offline and user-centric CRS metrics & Not a focus \\
\textbf{This review} & Dialogue, GUI, video, wearable, embodied, software engineering, human service & Decision pipeline $\times$ policy-construction mechanism & Instruction-relative, silence-gated decision under uncertainty & Common decision unit; triggering, timing, calibration, coverage--risk, off-policy value & Admissible set, tiered permission, recoverability \\
\bottomrule
\end{tabularx}
\end{table*}

\subsection{Paper Organization and Notation}

\begin{figure*}[t]
\centering
\includegraphics[width=\textwidth]{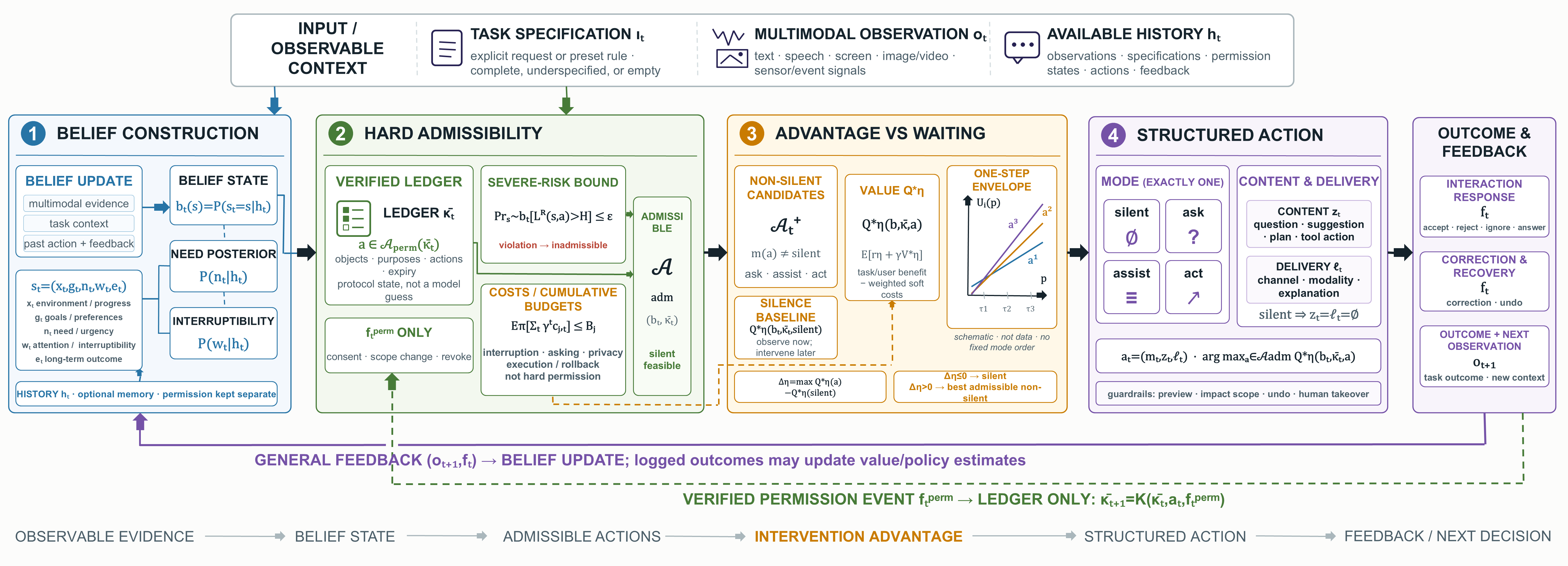}
\caption{Constrained proactive-service loop. The observable task specification \(\iota_t\) and verified permission ledger \(\bar\kappa_t\) remain separate from the latent-state belief \(b_t\). Within \(\mathcal A_{\mathrm{adm}}\), the agent compares intervention with the option value of silence and selects one structured action \(a_t=(m_t,z_t,\ell_t)\), where \(m_t\in\{\silent,\ask,\assist,\act\}\). The orange value envelope illustrates only the one-step preordered-candidate special case in Proposition~1; it neither imposes a fixed ordering on the four semantic modes nor represents empirical data. General feedback updates the belief, whereas only verified consent, scope-change, or revocation events update the permission ledger. Memory and personalization are optional state-construction mechanisms.}
\label{fig:unified}
\end{figure*}

Fig.~\ref{fig:unified} summarizes the survey. Section~II defines the problem and derives the gating structure. Section~III synthesizes methods along the decision loop. Section~IV examines benchmarks, metrics, and evidence. Section~V maps the same framework to three deployment regimes and states testable research questions. Section~VI concludes the paper.

The figure's symbols are defined formally in Section~II; here we note only the structure needed to read it. The observable task specification \(\iota_t\) and the verifiable permission ledger \(\bar\kappa_t\) are kept separate from the latent-state belief \(b_t\), whose state \(s_t=(x_t,g_t,n_t,w_t,e_t)\) collects the task environment, goals and preferences, need and urgency, interruptibility, and user-endorsed long-term outcome. The admissible set \(\mathcal A_{\mathrm{adm}}\) satisfies both authorization and the severe-risk bound, \(\mathcal A_t^+\) is its non-silent subset, and \(\Delta_{\boldsymbol\eta}\), evaluated under fixed nonnegative cost multipliers \(\boldsymbol\eta\), is the incremental value of the best admissible non-silent action over silence; the orange envelope illustrates only the one-step preordered-candidate special case of Proposition~1. General feedback \((o_{t+1},f_t)\) updates the belief, whereas only a verified permission event \(f_t^{\mathrm{perm}}\) updates the ledger through protocol operator \(K\).

\section{Problem Setting and Formal Foundations}

\subsection{Proactivity as a Decision Property Relative to an Instruction}

Let \(h_t\) denote the history available at time \(t\), and let \(\iota_t\) denote the observable task specification supplied by the current explicit request or a preset automation rule; authorization is represented separately by a verifiable ledger \(\bar\kappa_t\). The specification may completely determine an action, state only an underspecified objective, or be empty. The following definition separates a system speaking first from a system making a proactive decision.

\begin{definition}[Proactive service intervention]
A non-silent action \(a_t\) at time \(t\) is a \emph{proactive service intervention} relative to \(\iota_t\) if: (i) the system autonomously introduces at least one service opportunity, trigger time, service subgoal, information-acquisition action, or commitment to an external action not entailed by \(\iota_t\); (ii) the protocol permits silence, direct completion of the specified task, or another feasible action; (iii) the system selects the action from beliefs about the user's objective, the environment, and the consequences; and (iv) the objective includes user-endorsed benefit or shared task benefit. Choosing wording, a search path, or implementation details for already requested content is not itself a proactive intervention. A system that performs this gating in sequential interaction is a \emph{proactive service agent}.
\end{definition}

The definition is relative. A timer set in advance by the user speaks first, but its trigger is fully fixed by the rule and is therefore not a proactive decision in this sense. An agent that autonomously compares executing an underspecified request against clarifying it can be proactive. Task-oriented dialogue work operationalizes proactivity as introducing unrequested but goal-relevant information, while HRI work distinguishes anticipating a next step from taking initiative on the basis of that prediction \cite{investigatingproactivity2025,roman2024}. Complex planning, long-term memory, continuous sensing, and tool use are not necessary conditions; they enhance state estimation or action capability. Cross-disciplinary position work and information-seeking dialogue also use broader notions that emphasize timing, control, and trust or unsolicited additions to an answer. We use these workshop and position-preprint accounts only to delimit terminology, not as empirical validation of Definition~1 \cite{proactivesystems2026,redefining2024}. Table~\ref{tab:boundary} distinguishes adjacent concepts through initiative and online choice.

\begin{table*}[t]
\centering
\caption{Operational boundary between proactive service and adjacent interaction paradigms. The boundary depends on whether the trigger and action still require an online system choice, not on whether the interface speaks first.}
\label{tab:boundary}
\small
\setlength{\tabcolsep}{4.0pt}
\begin{tabularx}{\textwidth}{p{2.0cm}p{3.05cm}p{3.05cm}p{3.05cm}Y}
\toprule
Paradigm & Source of trigger or goal & Genuine alternatives at decision time & Relation to proactive service & Typical evidence \\
\midrule
Reactive agent & Current request determines the trigger and principal goal & Planning or execution within the given goal & Complex internal planning does not itself transfer initiative & Common LLM tool-use and GUI-agent setting \\
Preset automation & User or rule fixes the time and action beforehand & Usually no online silence gate & May present proactively, but does not make a proactive decision & Timed reminders and fixed workflows \\
Mixed initiative & Human and system may transfer control & Dialogue, action, or waiting & Proactive service is one class of system-initiated control transfer & \cite{walker1990,horvitz1999mixedinitiative} \\
Clarification & Current request has an information gap & Ask, act under an assumption, or decline & Can be epistemically proactive when asking changes the subsequent decision and is not protocol-mandated & \cite{aliannejadi2021clariq,rightquestion2024,coa2026} \\
Just-in-time adaptive intervention & Decision times and options are predefined; the rule varies with context & Intervene or not, and intervention content & A health-domain special case when it includes context-dependent gating & \cite{nahumshani2018jitai,lastjitai2025} \\
Autonomous agent & Degree of dependence on user involvement during execution & From advice to irreversible execution & Autonomy is orthogonal: a system may proactively advise without acting, or autonomously complete an explicit request & \cite{feng2025levelsautonomy} \\
\bottomrule
\end{tabularx}
\end{table*}

\subsection{Constrained Partially Observable Sequential Decision Making}

The central difficulty is that user goals, actual needs, interruptibility, and intended authorization are usually not directly observed, while intervention changes future states. Actual permission, by contrast, must be verified rather than inferred from behavior. We use a constrained partially observable Markov decision process as an \emph{analytical language}, without assuming that every existing system explicitly solves it \cite{kaelbling1998pomdp,poupart2015cpomdp}:
\begin{equation}
\mathcal M=\langle\mathcal S,\mathcal O,\mathcal A,T,Z,r,\mathbf c,\gamma,b_0\rangle .
\label{eq:cpomdp}
\end{equation}
Here, \(\mathcal M\) is the complete decision model; \(\mathcal S\), \(\mathcal O\), and \(\mathcal A\) are the latent-state, observation, and action spaces; \(T(s'\mid s,a)\) is the transition kernel from state \(s\) to next state \(s'\) after action \(a\); and \(Z(o,f\mid s',a,\bar\kappa)\) is the observation kernel that generates observation \(o\) and feedback \(f\) given the next state, action, and protocol state. The symbol \(r\) denotes the one-step reward function, \(\mathbf c=(c_1,\ldots,c_J)\) is the vector of \(J\) one-step cost functions, and \(b_0\in\Delta(\mathcal S)\) is the initial belief distribution, where \(\Delta(\mathcal S)\) is the set of probability distributions over \(\mathcal S\). We write \(r_t\) and \(c_{j,t}\) for the realized values of \(r\) and \(c_j\) on transition \(t\). The discount factor \(\gamma\in[0,1)\) makes rewards and costs \(t\) steps ahead decay geometrically through \(\gamma^t\): larger \(\gamma\) gives future states, long-term consequences, and the option value of waiting more influence on the current decision. Its reported value must therefore be interpreted together with the real duration represented by one decision step.
The latent state can be decomposed as
\begin{equation}
s_t=(x_t,g_t,n_t,w_t,e_t),
\label{eq:state}
\end{equation}
where \(x_t\) represents the environment and task progress, \(g_t\) the user's goals and preferences, \(n_t\) the latent need and urgency, \(w_t\) attention, workload, and interruptibility, and \(e_t\) user-endorsed long-term outcomes or relationship quality. The verifiable permission ledger \(\bar\kappa_t\) is protocol state, not a model guess: it records authorized objects, purposes, actions, and expiry. Latent willingness to grant permission may remain uncertain within \(g_t\), but cannot substitute for actual permission. The system receives multimodal observations \(o_t\). Acceptance, rejection, ignoring, an answer, correction, undo, and the task outcome after the preceding action are collectively feedback \(f_t\). The history and belief are
\begin{gather}
h_t=(o_0,\iota_0,\bar\kappa_0,a_0,f_0,\ldots,o_t,\iota_t,\bar\kappa_t),\\
b_t(s)=P(s_t=s\mid h_t),
\label{eq:belief}
\end{gather}
Thus, \(h_t\) chronologically aggregates observations, task specifications, ledger states, and previous actions and feedback through time \(t\). Under this indexing, \(f_t\) is observed after \(a_t\) and is used in the \(t\)-to-\(t+1\) update; before choosing \(a_t\), \(h_t\) contains past feedback only through \(f_{t-1}\). The symbol \(P\) is the probability operator, and \(b_t\in\Delta(\mathcal S)\) is the posterior belief over the latent state given that history; its role is to compress unobserved state uncertainty into a statistic suitable for decision making.
The ledger changes only through verifiable consent, scope-change, or revocation events:
\begin{equation}
\bar\kappa_{t+1}=K(\bar\kappa_t,a_t,f_t^{\mathrm{perm}}).
\label{eq:permissionupdate}
\end{equation}
Here, \(f_t^{\mathrm{perm}}\) is the subset of feedback \(f_t\) that the protocol has verified as a permission event, and \(K\) is the protocol transition operator that determines the next ledger from the current ledger, action, and verified event. Its role is to prevent ordinary acceptance, rejection, or behavioral cues from being treated as authorization.
Language expressing latent willingness changes the ledger only after the protocol verifies it as \(f_t^{\mathrm{perm}}\). The belief then updates after the action, feedback, new observation, and new ledger according to
\begin{equation}
b_{t+1}(s')\propto Z(o_{t+1},f_t\mid s',a_t,\bar\kappa_{t+1})
\sum_s T(s'\mid s,a_t)b_t(s).
\label{eq:update}
\end{equation}
In this update, \(s\) and \(s'\) are the current and next latent states, and the sum marginalizes the unknown current state. The kernels \(T\) and \(Z\) are those defined in Eq.~\eqref{eq:cpomdp}; \(\propto\) indicates normalization so that \(\sum_{s'}b_{t+1}(s')=1\). The equation updates belief from general feedback and a new observation; it does not update actual permission.

An action is a tuple rather than a single label:
\begin{equation}
a_t=(m_t,z_t,\ell_t),\qquad
m_t\in\{\silent,\ask,\assist,\act\},
\label{eq:action}
\end{equation}
where \(m_t\) is the intervention mode, \(z_t\) the concrete question, suggestion, plan, or tool action, and \(\ell_t\) the channel, modality, and explanation. By convention, \(z_t=\ell_t=\varnothing\) when \(m_t=\silent\). The \(\silent\) mode has no visible effect at the current time but permits further observation; \(\ask\) seeks a goal, evidence, preference, or authorization; \(\assist\) provides rejectable information, a suggestion, or a draft; and \(\act\) invokes tools, commits resources, or changes external state. Autonomy is encoded by the mode and the scope of \(z_t\), not by the delivery variable. Thus whether/when, what, and how are attributes of one action. Content and authority in turn change the gating threshold.

The one-step utility is
\begin{equation}
\begin{split}
r_t={}&B_t^{\mathrm{task}}+\beta B_t^{\mathrm{user,long}}
-\lambda_I C_t^{\mathrm{int}}-\lambda_Q C_t^{\mathrm{ask}}\\
&-\lambda_E C_t^{\mathrm{exec}}-\lambda_P C_t^{\mathrm{priv}},
\end{split}
\label{eq:reward}
\end{equation}
Here, \(B_t^{\mathrm{task}}\) is immediate task benefit and \(B_t^{\mathrm{user,long}}\) is a user-endorsed long-term benefit. The quantities \(C_t^{\mathrm{int}}\), \(C_t^{\mathrm{ask}}\), \(C_t^{\mathrm{exec}}\), and \(C_t^{\mathrm{priv}}\) are interruption, questioning, erroneous-execution or rollback, and privacy costs. The coefficient \(\beta\) scales long-term benefit relative to task benefit, while \(\lambda_I,\lambda_Q,\lambda_E,\lambda_P\geq0\) weight the respective costs. Their role is to map outcomes with different units into a comparable one-step net utility, covering task benefit, user-endorsed long-term outcomes, interruption, questioning burden, erroneous execution or rollback, and privacy cost. Trust, engagement, and acceptance are observations to be interpreted, not reward proxies that should automatically be maximized. The terms should either be scalarized with preregistered weights or be reported separately with their units and constraints. Classic mixed-initiative interfaces compare automated action, dialogue, and inaction by expected utility and explicitly model attention and interruptibility \cite{horvitz1999mixedinitiative,horvitz1999priorities,horvitz2003attention}. Equation~\eqref{eq:reward} extends that principle to structured actions of LLM agents.

Authorization cannot be only a soft penalty that sufficiently large utility can offset. Let \(\mathcal A_{\mathrm{perm}}(\bar\kappa_t)\) be the action set permitted by the verified ledger, with \(\silent\) always included, and let \(L_R(s,a)\) be a severe-risk loss. Let \(H\) be the threshold of unacceptable severe loss and \(\epsilon\in[0,1]\) the largest tolerated tail probability of exceeding it; \(\Pr_{s\sim b_t}\) takes probability over states drawn from belief \(b_t\). The admissible set under that belief is
\begin{equation}
\mathcal A_{\mathrm{adm}}(b_t,\bar\kappa_t)=
\left\{a\;\middle|\;
\begin{array}{l}
a\in\mathcal A_{\mathrm{perm}}(\bar\kappa_t),\\
\Pr_{s\sim b_t}[L_R(s,a)>H]\le\epsilon
\end{array}\right\}.
\label{eq:admissible}
\end{equation}
External execution is inadmissible when authorization is unknown, although an \(\ask\) action that requests authorization can remain admissible. The proactive gate below is restricted to decision opportunities for which \(\silent\in\mathcal A_{\mathrm{adm}}\). A protocol-mandated escalation has no genuine silence alternative and falls outside Definition~1 at that moment. Let \(c_{j,t}\) be the one-step interruption, privacy, or questioning cost of type \(j\), \(B_j\) the corresponding user or institutional budget, and \(J\) the number of cost types. A policy \(\pi(a\mid b,\bar\kappa)\) maps the current belief and ledger to an action distribution, while \(\mathbb E_\pi\) takes expectation under the trajectory distribution jointly induced by \(\pi\), \(T\), and \(Z\). The optimal policy satisfies
\begin{equation}
\begin{aligned}
\pi^*&=\arg\max_{\pi}\mathbb E_{\pi}
\left[\sum_{t=0}^{\infty}\gamma^t r_t\right],\\
\text{s.t.}\quad
\mathbb E_{\pi}\!\left[\sum_{t=0}^{\infty}\gamma^t c_{j,t}\right]&\le B_j,
\quad j=1,\ldots,J,\\
a_t&\in\mathcal A_{\mathrm{adm}}(b_t,\bar\kappa_t)
\quad\text{a.s. for all }t .
\end{aligned}
\label{eq:policy}
\end{equation}
The \(\arg\max\) selects the policy with the largest discounted cumulative reward, and \(\pi^*\) is optimal subject to all cumulative budgets and per-step admissibility. The abbreviation \(\mathrm{a.s.}\) means ``almost surely'': except on probability-zero trajectories, every selected action must be admissible at every time. The same \(\gamma\) defined in Eq.~\eqref{eq:cpomdp} controls the relative weight of future rewards and future costs.
This formulation distinguishes an expensive but selectable action from one that must not be executed without consent. To close the action value under these cumulative budgets, we use a Lagrangian relaxation with fixed nonnegative multipliers \(\boldsymbol\eta=(\eta_1,\ldots,\eta_J)\), define the one-step cost vector \(\mathbf c_t=(c_{1,t},\ldots,c_{J,t})\), and set \(r_{\boldsymbol\eta,t}=r_t-\boldsymbol\eta^\top\mathbf c_t\). Here \(\boldsymbol\eta^\top\mathbf c_t\) is the inner product of the multiplier and cost vectors, and each \(\eta_j\) is the shadow price of budget constraint \(j\), incorporating that soft cost into the relaxed reward. The function \(r_{\boldsymbol\eta}(s,a,s')\) denotes the corresponding relaxed reward function, and \(r_{\boldsymbol\eta,t}\) is its realization on transition \(t\). Dual optimization or budget calibration can select the multipliers. If strong duality or feasibility conditions fail, the relaxed gate only yields a candidate policy whose cumulative budgets must be checked separately. The following \(Q^*_{\boldsymbol\eta}\), \(V^*_{\boldsymbol\eta}\), and \(\Delta_{\boldsymbol\eta}\) refer to this fixed-multiplier relaxed problem rather than an unconstrained value function.

\subsection{Intervention Advantage, the Value of Waiting, and the Value of Asking}

Define the optimal action value at a belief state as
\begin{equation}
\begin{aligned}
Q^*_{\boldsymbol\eta}(b,\bar\kappa,a)
={}&\mathbb E\big[r_{\boldsymbol\eta}(s,a,s')\\[-.3ex]
&+\gamma V^*_{\boldsymbol\eta}(b',\bar\kappa')\mid b,\bar\kappa,a\big].
\end{aligned}
\label{eq:q}
\end{equation}
Here, \(Q^*_{\boldsymbol\eta}(b,\bar\kappa,a)\) is the discounted value of first taking action \(a\) under belief \(b\) and ledger \(\bar\kappa\), then acting optimally; \(V^*_{\boldsymbol\eta}\) is the corresponding optimal state value, and the superscript \(^*\) denotes optimization over subsequent policies. The variables \(b'\) and \(\bar\kappa'\) are the belief and ledger after one transition, \(r_{\boldsymbol\eta}(s,a,s')\) is its Lagrangian-relaxed one-step reward, and the conditional expectation averages over uncertainty in the next state, observation, and feedback.
Let \(\mathcal A_t^+=\{a\in\mathcal A_{\mathrm{adm}}(b_t,\bar\kappa_t):m(a)\neq\silent\}\), where \(m(a)\) returns the mode component of the action tuple and the \(+\) only marks the non-silent admissible subset rather than positivity. The incremental value of intervention relative to silence is
\begin{equation}
\begin{aligned}
\Delta_{\boldsymbol\eta}(b_t,\bar\kappa_t)
={}&\max_{a\in\mathcal A_t^+}Q^*_{\boldsymbol\eta}(b_t,\bar\kappa_t,a)\\[-.3ex]
&-Q^*_{\boldsymbol\eta}(b_t,\bar\kappa_t,\silent).
\end{aligned}
\label{eq:advantage}
\end{equation}
If \(\mathcal A_t^+=\varnothing\), the maximum in Eq.~\eqref{eq:advantage} is defined as \(-\infty\). We use conservative tie-breaking: an optimal agent remains silent when \(\Delta_{\boldsymbol\eta}(b_t,\bar\kappa_t)\leq0\) and otherwise selects the admissible action with highest value. Here \(Q^*_{\boldsymbol\eta}(b,\bar\kappa,\silent)\) includes the option to observe further and intervene later, so silence is not permanent abandonment. For a stochastic policy \(\pi\), let \(a_t\sim\pi(\cdot\mid b_t,\bar\kappa_t)\). Its first intervention time is
\begin{equation}
T_{\mathrm{int}}^{\pi}=\inf\{t:m(a_t)\neq\silent\},
\quad a_t\sim\pi(\cdot\mid b_t,\bar\kappa_t).
\label{eq:stopping}
\end{equation}
The infimum \(\inf\) selects the earliest time satisfying the condition; if no non-silent action ever occurs, \(T_{\mathrm{int}}^\pi=+\infty\) by convention. This random stopping time describes when the policy actually first intervenes rather than a trigger fixed in advance.
For the relaxed optimal policy under fixed \(\boldsymbol\eta\), conservative tie-breaking makes this equal to the first time of positive intervention advantage along a trajectory that has remained silent. This also explains why predicting whether a need currently exists is generally insufficient. With the same need probability, poor candidate content or high execution risk can produce a different intervention advantage.

The total value of asking includes its immediate interaction effect and the way an answer changes later decisions. In the special case where an answer returns immediately and \(y\) is a sufficient statistic for the resulting observation and feedback, question \(q\) has value
\begin{equation}
\begin{aligned}
Q^*_{\boldsymbol\eta}(\ask(q)\mid b,\bar\kappa)
={}&\mathbb E_{y\sim P(y\mid b,\bar\kappa,q)}\!\left[
r_{\boldsymbol\eta}(b,\bar\kappa,\ask(q),y)\right.\\
&\left.{}+\gamma V^*_{\boldsymbol\eta}(b^{q,y},\bar\kappa^{q,y})\right].
\end{aligned}
\label{eq:voi}
\end{equation}
Here, \(\ask(q)\) is an asking action whose content is question \(q\), \(P(y\mid b,\bar\kappa,q)\) is the predictive distribution of possible answer \(y\), \(b^{q,y}\) is the posterior belief after that answer, and \(\bar\kappa^{q,y}\) is the subsequent ledger, which changes only if the answer contains a verified permission event. The term \(r_{\boldsymbol\eta}(b,\bar\kappa,\ask(q),y)\) is the conditional expected immediate relaxed reward of asking given the belief, ledger, and received answer \(y\). The expectation averages over all possible answers, adding this immediate reward to the discounted optimal value after the answer.
Only a protocol-verified consent or revocation event may change \(\bar\kappa\). In a one-step approximation that ignores other immediate effects, let \(V_0\) denote the optimal baseline value when the current question is disallowed and the process proceeds directly to the next decision, and let \(C_Q(q)\) be the direct burden or cost of asking \(q\). The pure value of information is then \(\mathbb E_y[V_0(b^{q,y},\bar\kappa^{q,y})]-V_0(b,\bar\kappa)-C_Q(q)\). If an answer changes neither the belief, authorization, nor downstream action values, this pure information value cannot be positive. A question may nevertheless have an immediate politeness, commitment, or relational effect. Qulac and clarifying-question generation first made whether and what to ask directly evaluable in information retrieval, while conversational recommendation later used questions about usage context for preference elicitation \cite{aliannejadi2019qulac,zamani2020clarifyingq,balog2024}. Recent clarification work further learns gating and question construction jointly \cite{rightquestion2024,stargate2024,doubleturn2025,coa2026,askorassume2026}. Visual ambiguity likewise changes the value of asking: AQuA selects among inference, enumeration, and clarification according to ambiguity, whereas the FoRLM workshop paper ClarifyVQA studies question generation under missing context. The latter is workshop-level method evidence, not evidence of deployment benefit \cite{aqua2026,clarifyvqa2025}.

\begin{proposition}[One-step threshold special case for preordered candidates]
Fix the other state variables and authorization, let the latent need be \(N\in\{0,1\}\) and \(p=P(N=1\mid b)\in[0,1]\), and let an application preorder a finite candidate set by interference or risk as \(a^{(1)}\prec\cdots\prec a^{(K)}\). If \(U_i(p)=u_{i0}+p d_i\), the optimal utility is the upper envelope of these affine functions. For any pair with \(d_i\neq d_j\), a valid switch can occur only at an intersection in \([0,1]\):
\begin{equation}
p_{ij}=\frac{u_{i0}-u_{j0}}{d_j-d_i}.
\label{eq:threshold}
\end{equation}
Here, \(K\) is the number of candidate actions, \(u_{i0}=U_i(0)\) is candidate \(i\)'s utility intercept when the need variable is zero, \(d_i\) is its utility slope with respect to need probability \(p\), and \(p_{ij}\) is a potential switch probability at which candidates \(i\) and \(j\) have equal utility. Only an intersection in \([0,1]\) that lies on the upper envelope changes the optimal action.
If \(d_i\) is nondecreasing in the predefined order, then \(U_j(p)-U_i(p)\) is nondecreasing in \(p\) for every \(j>i\). Under fixed tie-breaking, the index of the optimal action is therefore nondecreasing in \(p\). Candidates dominated on the upper envelope may disappear from the optimal policy, and collinear candidates are resolved by the tie-breaking rule.
\end{proposition}

The proof follows because the upper envelope of affine functions changes only where two functions are equal, and the single-crossing condition prevents pairwise preference from reversing twice as \(p\) increases. The classic mixed-initiative expected-utility analysis gives one instance with inaction, dialogue, and automated execution \cite{horvitz1999mixedinitiative}; the proposition does not assume that \(\ask,\assist,\act\) naturally form a linear autonomy order. Lowering an intercept moves a threshold according to Eq.~\eqref{eq:threshold} only when the added interruption or risk cost is independent of \(N\) and the remaining terms are fixed. In general, a gate based only on need probability and a fixed threshold is not guaranteed to be optimal unless incremental value is monotone in that probability and other action attributes are fixed or correctly marginalized.

\section{A Decision-Centered Synthesis of Methods}

\begin{figure*}[t]
\centering
\includegraphics[width=\textwidth]{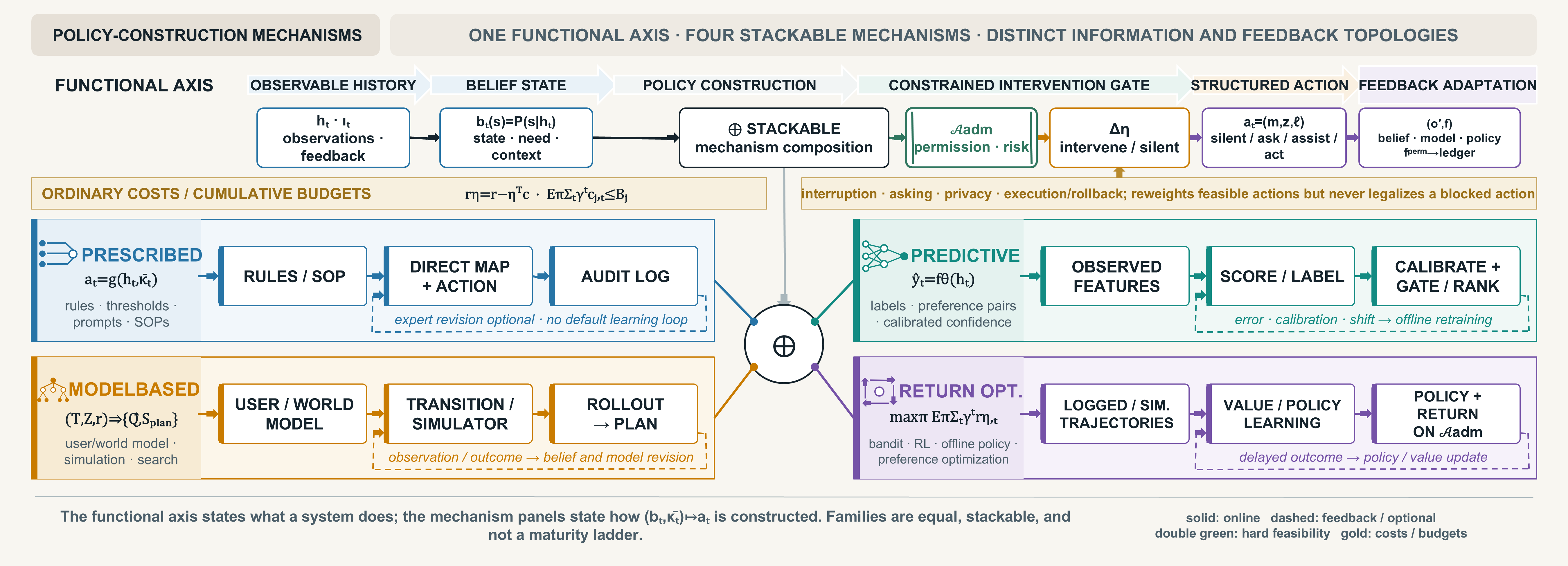}
\caption{Decision-centered functional axis and non-exclusive policy-construction mechanisms. A complete system maps history to a belief state, gates intervention, constructs a structured action, and adapts from feedback. Prescribed, predictive, model-based, and return-optimizing mechanisms may span and combine across modules; row formulas and descriptors are mechanism signatures rather than empirical coverage. Verified permission and severe-risk constraints define hard feasibility; interruption, asking, privacy, and execution or rollback enter costs or budgets instead of being treated as equivalent hard constraints.}
\label{fig:taxonomy}
\end{figure*}

Fig.~\ref{fig:taxonomy} gives the functional axis used in this survey. A complete system converts history into a belief state, gates intervention within the feasible set, constructs and executes an action, and uses feedback to update its state or policy. Sensing modality, application domain, autonomy level, and training paradigm are orthogonal properties rather than classes parallel to these four modules. Representative systems often cover only part of the loop; recording that coverage helps distinguish proactive decision mechanisms from general supporting technology.

The four mechanism signatures in the figure use the following notation. The map \(g\) is directly specified by rules, thresholds, prompts, or SOPs. The predictor \(f_{\theta}\) has parameters \(\theta\), and \(\hat y_t\) is its estimated need, gate, or action label/score, with a hat denoting an estimate. Model-based methods explicitly use transition, observation, and reward components \((T,Z,r)\); \(\hat Q\) and \(S_{\mathrm{plan}}\) denote an estimated action value and a planning/search state. Finally, \(\max_{\pi}\mathbb E_{\pi}\sum_t\gamma^t r_{\boldsymbol\eta,t}\) denotes maximization of discounted cumulative relaxed reward over policies \(\pi\). These signatures identify how the policy mapping is constructed; they do not rank the four mechanisms.

\subsection{From Observation to a Decision-Ready Belief State}

\textbf{Opportunity detection and temporal segmentation.} In streaming screens, egocentric video, or sensor input, most moments contain no service event. A system must first segment observations into decision opportunities and distinguish task signals from background activity and noise. PIRA-Bench evaluates interleaved latent intents in manually curated sequential GUI-screenshot trajectories \cite{pirabench2026}; ESTP-Bench couples answer content with an appropriate response time in egocentric video \cite{eyeswideopen2025}; ContextAgent and Alpha-Service map multimodal context to tool or service opportunities \cite{contextagent2025,alphaservice2025}. AppAgent-Pro, published at CIKM 2025, further integrates cross-application information to extend a literal request, illustrating how state construction can surface latent subgoals; it does not establish an open-stream silence gate \cite{appagentpro2025}. In these streaming resources, key errors beyond language generation include omissions, duplicate segments, and noncausal use of future frames. Evaluation should retain a predefined complete negative stream and timestamps.

\textbf{Need, interruptibility, and uncertainty.} Moving from an event to action also requires estimating whether help is needed, its urgency, and whether the user can be interrupted. PASK combines need detection with user, workspace, and global memory \cite{pask2026}; Satori uses a belief--desire--intention model to explain the next need \cite{satori2025}; and ProMemAssist modulates intervention timing through working-memory and interference models \cite{promemassist2025}. The unified framework requires calibrated uncertainty rather than only an intent label, because Eqs.~\eqref{eq:admissible} and \eqref{eq:advantage} respectively evaluate a tail-risk probability and the expected value of waiting.

\textbf{The conditional role of memory.} Long-term memory matters for cross-session preferences, recurrent tasks, and personal routines. MemoryOS, HyMEM, and PersonalAlign respectively investigate hierarchical storage, structured trajectory memory, and personalization from longitudinal records \cite{memoryos2025,hymem2026,lyu2026personalalign}. Memory nevertheless changes only how \(b_t\) is constructed: a warning based on an immediate hazard may use no long-term memory, whereas a system retrieving years of history only after an instruction remains reactive. Longer memory also creates stale-information, subject-mismatch, and privacy costs. Provenance, confidence, retention period, and deletion rights should therefore be represented in the state.

\subsection{How Policies Are Constructed: Four Non-Exclusive Mechanisms}

Table~\ref{tab:methods} gives four stackable labels for how the online mapping \((b_t,\bar\kappa_t)\mapsto a_t\) is constructed: whether a rule directly specifies the policy, a model fits need or action labels, transition or observation dynamics are explicit, and sequential return is optimized directly. A compound system may receive several labels. For example, offline reinforcement learning with Monte Carlo tree search contains both return-optimizing and model-based components, while whether its gate includes \(\silent\) still requires a separate annotation.

\begin{table*}[t]
\centering
\caption{Four non-exclusive mechanisms for constructing a proactive policy. Representative studies illustrate components; their inclusion does not imply that every study implements the complete four-mode gate.}
\label{tab:methods}
\small
\setlength{\tabcolsep}{3.7pt}
\begin{tabularx}{\textwidth}{p{1.8cm}p{3.1cm}p{2.9cm}p{2.85cm}Y}
\toprule
Label & Online decision form & Main design or training signal & Advantages and suitable conditions & Typical failures and representative work \\
\midrule
Prescribed & Rules, thresholds, prompts, or SOPs directly specify an action or policy & Expert rules, few-shot examples, program constraints & Transparent and inexpensive; suitable for stable permissions and enumerable risks & Thresholds transfer poorly across users and long-horizon effects remain implicit; prompts and SOPs in \cite{deng2023procot,controllablemixedinit2023,li2025chatsop} \\
Predictive & A supervised model directly estimates a need, gate, or action label & Human/model labels, preference pairs, confidence & Scales to fixed protocols and supports high-throughput offline testing & Labels compress utility into a point target and calibration degrades under shift; examples in \cite{lu2024proactiveagent,pask2026,rightquestion2024,askorassume2026} \\
Model-based & Explicit user/environment model, Bayesian update, POMDP, BDI, or search & Transition and observation models, simulator, heuristic value & Compares waiting, asking, and future effects, and makes constraints interpretable & Model mismatch and planning cost; examples in \cite{horvitz1999mixedinitiative,bayesianproactive2024,satori2025,whennottohelp2025,he2024dpdp,zhang2024pcqpr} \\
Return-optimized & Bandits, RL, offline policy learning, or preference optimization maximize cumulative return & Outcome/process rewards, user or simulated feedback & Handles delayed outcomes and coupling between actions & Reward misspecification, simulator bias, and insufficient offline support; examples in \cite{sotopiarl2025,userrl2025,itpo2026,togate2026,speakrl2026,clariti2026} \\
\bottomrule
\end{tabularx}
\end{table*}

Prescribed and predictive components commonly compress future effects into expert rules or labels. They are computationally light but have difficulty expressing the option value of waiting. Model-based components explicitly expand \(T,Z\), or a search tree, and can evaluate Eq.~\eqref{eq:q}, but are sensitive to user-model error. Return-optimizing components estimate long-horizon value directly, yet if their training environment lacks silence, rejection, or authorization violations, they still optimize only a content policy. In multi-turn strategy tasks where a speaking turn is already available, EPO and CSO improve policy sequences through explicit strategic reasoning and preference/process optimization, respectively. They show that return optimization can act on content \(z_t\), not that proactive gating over mode \(m_t\) has been learned \cite{epo2025,cso2025}. Component comparisons therefore cannot be separated from the action space and feedback protocol.

\subsection{Joint Intervention Gating and Action Construction}

The gate selects a mode and time; the content policy selects a specific question, suggestion, or tool plan; and the delivery policy chooses channel, modality, and explanation. Autonomy and external side effects are represented by mode and permission. Goal guidance and policy planning in proactive dialogue \cite{wu2019duconv,tang2019targetguided,topdial2023,ppdpp2024}, preference elicitation and goal planning in conversational recommendation \cite{liu2020durecdial,liu2023mgcg,wang2022followme,tpra2025,ipg2024}, and clarification in software engineering \cite{ambigswe2025,clariti2026} provide content-construction mechanisms, but many listed experiments already grant the system a speaking turn. DuRecDial~2.0, TG-ReDial, and INSPIRED add bilingual, topic-guided, and sociable recommendation resources, but their protocols largely condition on an ongoing dialogue; they therefore support action construction and cross-lingual evaluation rather than the silence gate \cite{liu2021durecdial2,tgredial2020,inspired2020}. A WWW 2025 companion paper on influence-path planning and a separate preprint on joint timing--content recommendation provide emerging mechanism evidence, not evidence of long-term deployment benefit \cite{influencepath2024,pasrec2025}. Applying these methods to open-world proactive service also requires estimating the conditional value of content: even a high-quality answer can have negative net utility at the wrong time. A notification-optimization preprint jointly models whether to send and future response, furnishing a methodological example of comparing intervention with silence over a longer horizon; because no formal proceedings record was verified, we do not treat it as published evidence \cite{pushnotif2022}.

The \(\ask\), \(\assist\), and \(\act\) modes are not a simple linear scale of autonomy. Asking can reduce epistemic uncertainty but adds cognitive burden; assistance preserves user control but can create choice architecture and overreliance; execution can save effort but introduces permission, rollback, and accountability. Search and planning methods such as DPDP, PCQPR, EPL, and ChatSOP compare downstream effects of candidate policies \cite{he2024dpdp,zhang2024pcqpr,epl2024,li2025chatsop}. To implement proactive gating in the sense of Eq.~\eqref{eq:advantage}, their root candidates should also contain \(\silent\) and the authorization levels feasible at that time. Content evaluation should separately report conditional quality under a reference gate and end-to-end quality under the system gate, so that upstream false interventions remain visible.

\subsection{Feedback, Personalization, and Continual Adaptation}

Feedback operates at least at three timescales. Immediate feedback (an answer, rejection, or undo) updates the current belief. Task outcomes update the value of content and execution. Across sessions, acceptance, trust, and burden change the long-term user model. These representative studies refresh memory or retrieve prior experience but do not provide controlled online updates of the gating policy \cite{epl2024,memoryos2025,hymem2026,lyu2026personalalign}. These operations should be distinguished: writing a new event into memory does not show that a model maintains a stability--plasticity balance under preference drift \cite{wang2024continual}.

Acceptance or task outcomes observed only after intervention create selective feedback. The system sees a rejection after a poor suggestion, but when it remains silent it cannot observe whether help was needed. Directly using visible acceptance as reward favors moments that are easy to accept and may overlook potential beneficiaries. Continual adaptation generally requires versioned regression sets and preserved authorization boundaries. Offline policy learning and counterfactual evaluation additionally require logging-policy probabilities, and exploration is appropriate only when ethical and safety conditions permit it. Section~IV accordingly evaluates classification, timing, and policy value separately.

\section{Evaluation Protocols, Benchmarks, and Evidence}

\subsection{A Common Decision Unit and Benchmark Schema}

A comparable proactive-service example should use a \textbf{predefined complete stream of decision opportunities} as its unit, rather than collecting only positive cases in which the system should help. A study should state whether opportunities are constructed from fixed time slices, event changes, task stages, or user-acceptable windows. A minimal record includes session, user, and time; visible observations and the history window; current request status; a service opportunity and its valid interval; permissible actions; authorization and risk; the selected action and confidence; logging-policy probability; whether the user saw, accepted, rejected, or undid the action; and task and long-term outcomes. Label provenance (real user, expert, simulator, or LLM) should also be explicit. Conversation-shape work shows that initiative and collaboration structure can be operationalized, although such dialogue-level measures are not deployment utility \cite{conversationshape2020}; multimodal ProactiveBench adds proactive inquiry under incomplete visual evidence \cite{demin2025proactivebench}. Table~\ref{tab:benchmarks} restates representative resources using the variables above, rather than comparing event counts, stages, and dataset counts in a single scale column. To avoid collapsing realism, comparison, and follow-up into one rank, R0--R4 describes only interaction/deployment realism from constructed cases to in-situ use; C0/C1 records whether a prespecified condition or policy comparison exists; and H-NA/H-S/H-L denotes no human outcome, short-session outcomes, or longitudinal follow-up linked at the individual level. The descriptors are independent, and C1 does not by itself establish causality.

\begin{table*}[t]
\centering
\caption{Representative proactive-service resources and studies restated under a common decision protocol. Modes are only those directly supervised or evaluated; R/C/H denote interaction realism, comparison design, and human-outcome horizon.}
\label{tab:benchmarks}
\footnotesize
\setlength{\tabcolsep}{3.0pt}
\renewcommand{\arraystretch}{1.04}
\begin{tabularx}{\textwidth}{@{}p{0.115\textwidth}p{0.285\textwidth}p{0.105\textwidth}Yp{0.095\textwidth}@{}}
\toprule
Resource & Decision unit and source & Direct modes & Protocol and direct outcome & R/C/H \\
\midrule
ClariQ \cite{aliannejadi2021clariq} & Queries/topics; human clarification and relevance labels & \silent/\ask & Static need classification and question ranking/retrieval & R0/C0/H-NA \\
ProactiveBench \cite{lu2024proactiveagent} & Text events; 6,790 synthetic train, 233 real test & \silent/\assist & Discrete positives/negatives; P/R/F1 and reward-model acceptance proxy & R0/C0/H-NA \\
PROBE \cite{probe2025} & 1,000 synthetic long-document professional scenarios & \assist & Injected bottleneck in every case; retrieval, identification, and plan selection & R0/C0/H-NA \\
Ambig-SWE \cite{ambigswe2025} & Underspecified coding tasks; Full/Hidden/Interaction conditions & \ask/\act & Simulated interaction; detection, question quality, and solve rate & R2/C1/H-NA \\
ESTP-Bench \cite{eyeswideopen2025} & 890 Ego4D videos; 2,264 human-verified timed QA & \silent/\assist & Sequential frames and answer windows; ESTP-F1, latency/FPS & R1/C0/H-NA \\
PIRA-Bench \cite{pirabench2026} & 100 curated GUI screenshot trajectories; injected/pure noise & \silent/\assist & Complete offline stream; mean F1 and normalized false alarms & R1/C0/H-NA \\
LatentNeeds-Bench \cite{pask2026} & 100 real transcribed sessions (3,936 turns) & \silent/\assist & Multiturn binary need; balanced accuracy and latency & R1/C0/H-NA \\
ProAgentBench \cite{proagentbench2026} & One-month computer logs; LLM invocation as need proxy & \silent/\assist & Matched non-invocation negatives; timing, intent, and query similarity & R1/C0/H-NA \\
ProactiveEval \cite{liu2026proactiveeval} & Six simulated multiturn text scenarios & \ask/\assist & Staged interaction; planning/guidance rubric & R2/C0/H-NA \\
ProactBench \cite{proactbench2026} & Planner-generated history and fixed trigger turn & \assist & No silence gate; stage and overall response scores & R0/C0/H-NA \\
When Not to Help \cite{whennottohelp2025} & Three policies in a simplified POMDP user model & \silent/\assist & Long-horizon simulation; model-internal return/engagement & R2/C1/H-NA \\
ProMemAssist \cite{promemassist2025} & 12 participants; four glasses/object tasks; within-subject study & \silent/\assist & About 60 minutes; messages, reactions, scales, and burden & R3/C1/H-S \\
SCALA \cite{scala2026} & 1,508 students; semester-long in-situ classroom deployment & \assist & 2,608 sessions; anonymous session-level feedback & R4/C0/H-S \\
\bottomrule
\end{tabularx}
\end{table*}

The table exposes three gaps. First, R0 resources diagnose static capabilities but cannot estimate how intervention changes a later trajectory. Every PROBE case contains a bottleneck, so it tests what plan to identify and select rather than when to remain silent in an open stream; ProactBench likewise regenerates a response only at a fixed trigger turn. Second, R1 complete streams improve negative and timing evaluation but remain offline replays whose observations are unaffected by model actions. ProAgentBench uses the time of an LLM invocation as a proxy for need rather than user-confirmed counterfactual ground truth. Third, R2/C1 simulation supports policy comparison only inside its user model; ProMemAssist provides a short designed human comparison, whereas SCALA provides semester-long observational deployment but, by design, does not link student identities to their queries and therefore cannot provide individual longitudinal outcomes. The listed studies reach R4, but not H-L, and none uses a comparison design that identifies long-term incremental intervention value. Offline scores therefore do not establish long-term net utility.

\begin{figure*}[t]
\centering
\includegraphics[width=\textwidth]{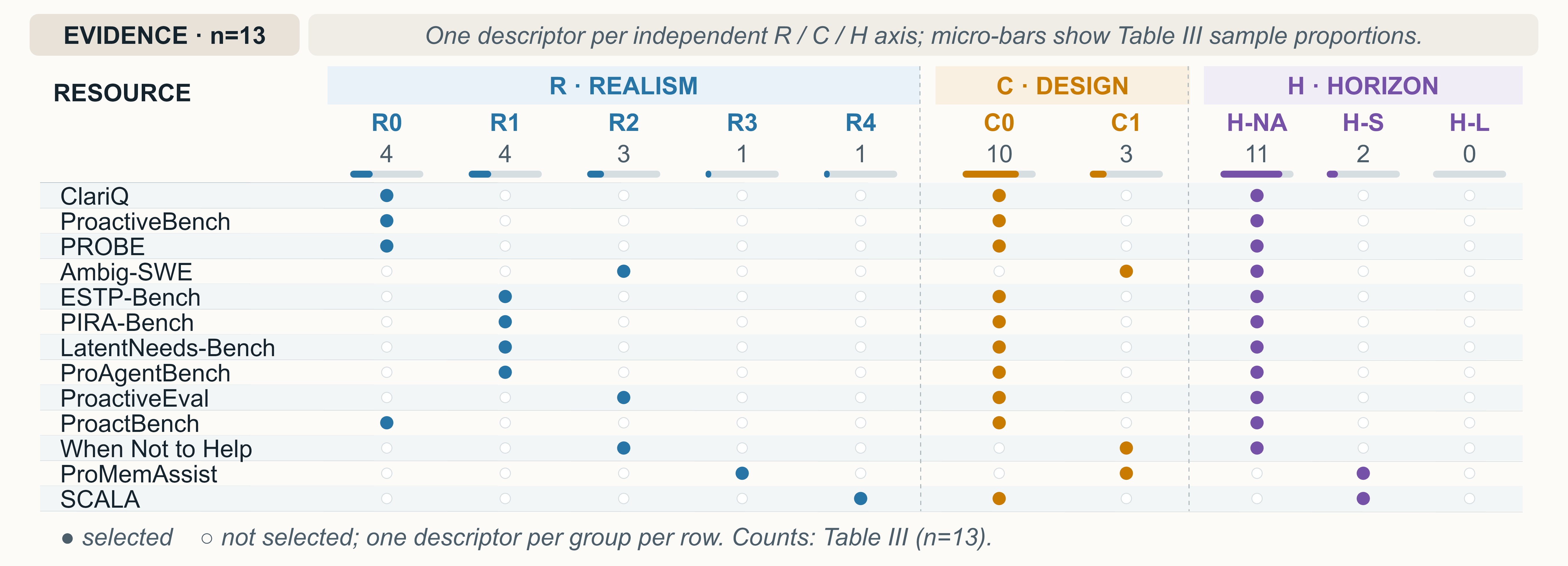}
\caption{Per-resource evidence matrix for the 13 representative resources and studies in Table~III. Each row records one interaction-realism descriptor (R0--R4), one comparison descriptor (C0/C1), and one human-outcome-horizon descriptor (H-NA/H-S/H-L); numbers above columns are marginal counts. R, C, and H remain independent, and the matrix describes only the table sample rather than field prevalence. Higher R or C1 alone does not establish causal identification or longitudinal benefit.}
\label{fig:evidence}
\end{figure*}

\subsection{From Classification Scores to Policy Value}

\textbf{Gating and calibration.} For \(N\) decision opportunities with need labels \(y_i\in\{0,1\}\), the prevalence \(\rho=N^{-1}\sum_i y_i\) must be reported first. Accuracy in a sparse stream can be achieved by always remaining silent. Evaluation should therefore combine precision, recall, F1 or PR-AUC, balanced accuracy, and false interventions per hour or session:
\begin{equation}
\mathrm{FAH}=\frac{\#\{\text{false interventions}\}}
{\text{observed hours}}.
\label{eq:fah}
\end{equation}
Here, \(\#\{\cdot\}\) is an event count: the numerator counts interventions without a reference need and the denominator is the total duration under complete observation. Thus \(\mathrm{FAH}\) is the false-intervention rate per observed hour. In this paragraph, \(N\) is the number of decision opportunities, \(i\) indexes an opportunity, \(y_i\) is its binary reference label, \(\rho\) is prevalence, and \(\hat p_i\) is the estimated need probability.
If the system outputs a need probability \(\hat p_i\), the Brier score \(N^{-1}\sum_i(\hat p_i-y_i)^2\), calibration curves, and expected calibration error with confidence intervals should accompany discrimination metrics. ECE is sensitive to binning and cannot be the sole calibration evidence.

\textbf{Event-level timing.} Let the reference service events be \(\mathcal E=\{([l_j,u_j],t_j^*)\}\) and the predicted intervention times be \(\mathcal P\). For event \(j\), \([l_j,u_j]\) is its valid intervention interval, with lower and upper bounds \(l_j\) and \(u_j\), and \(t_j^*\) is the reference ideal time inside that interval. Form a bipartite graph in which a prediction connects to an event when it falls in the valid interval. Select a maximum-cardinality one-to-one matching, and, if several exist, the one with minimum total absolute timing error; denote it by \(M\subseteq\mathcal P\times\mathcal E\). Then
\begin{equation}
\mathrm{Prec}_T=\frac{|M|}{|\mathcal P|},\quad
\mathrm{Rec}_T=\frac{|M|}{|\mathcal E|},\quad
F_{1,T}=\frac{2\mathrm{Prec}_T\mathrm{Rec}_T}{\mathrm{Prec}_T+\mathrm{Rec}_T}.
\label{eq:timingf1}
\end{equation}
Here, \(|\cdot|\) is set cardinality, and \(\mathrm{Prec}_T\), \(\mathrm{Rec}_T\), and \(F_{1,T}\) are event-level timing precision, recall, and their harmonic mean; subscript \(T\) distinguishes timing from content quality.
If a corresponding denominator is empty, the metric is N/A rather than zero; if \(\mathrm{Prec}_T+\mathrm{Rec}_T=0\), \(F_{1,T}\) is likewise recorded as N/A under the preregistered convention. For every matched event, the signed error \(e_j=\hat t_j-t_j^*\) should be reported, where \(\hat t_j\) is the prediction matched to reference event \(j\); \(e_j<0\) and \(e_j>0\) indicate early and late intervention. Delay to the first valid intervention should be summarized separately. Repeated alerts for one event match only once. A fixed frame tolerance that is unrelated to a user-acceptable interval replaces utility with annotation convenience.

\textbf{Action and end-to-end decomposition.} Four-mode selection can be evaluated with macro-F1 or a cost-weighted confusion matrix, whereas content can use solve rate, verifiable correctness, or human judgment as appropriate. Every content metric should include at least two conditions: evaluate \((z,\ell)\) under reference trigger conditions and evaluate the end-to-end system under its own gate. The first localizes action-construction capability; the second captures error propagation from false triggers. Acceptance, ignoring, rejection, correction, and undo should also be separated. A high acceptance rate can result from selecting easy users or persuasive defaults and cannot replace net utility.

\textbf{Policy and causal value.} Let \(Y^\pi\) be the task or long-term outcome under policy \(\pi\), and let \(\mathbf C^\pi\) be a vector of user burden, privacy, and risk costs. With preregistered, dimensionally explicit scalarization weights \(\boldsymbol\lambda\), deployment value relative to silence or an existing system \(\pi_0\) is
\begin{equation}
\Delta V(\pi,\pi_0)=
\mathbb E[Y^\pi-Y^{\pi_0}]-
\boldsymbol\lambda^\top\mathbb E[\mathbf C^\pi-\mathbf C^{\pi_0}].
\label{eq:policyvalue}
\end{equation}
Here, \(\mathbb E\) averages over user and environment outcomes induced by the corresponding policy, \(\boldsymbol\lambda^\top\mathbf C\) is the inner product of the cost vector and nonnegative scalarization weights, and \(\Delta V(\pi,\pi_0)\) is the target policy's net incremental value over the baseline. A positive value means that incremental benefit exceeds added cost under the declared weights.
The same trajectory cannot reveal outcomes both with and without intervention. In the contextual-bandit special case (one decision opportunity followed by its outcome), data logged by a behavior policy \(\mu\) with support overlap admit the inverse propensity estimator
\begin{equation}
\widehat V_{\mathrm{IPS}}(\pi)=\frac1N\sum_{i=1}^{N}
\frac{\pi(a_i\mid h_i)}{\mu(a_i\mid h_i)}R_i,
\label{eq:ips}
\end{equation}
In Eq.~\eqref{eq:ips}, \(N\) is the number of independent logged decision units and \(i\) indexes a unit; \(h_i\), \(a_i\), and \(R_i\) are its logged history, action, and realized return. The probabilities \(\mu(a_i\mid h_i)\) and \(\pi(a_i\mid h_i)\) are the propensities of that action under the behavior and target policies, and their ratio reweights the behavior log toward the target-policy distribution. The hat on \(\widehat V_{\mathrm{IPS}}\) denotes an estimate from finite data. Support overlap requires every action with positive target-policy probability to have positive behavior-policy probability.
However, proactive interventions usually change subsequent states, so a whole trajectory cannot be compressed into independent samples. For sequential trajectory \(i\), define
\begin{equation}
\begin{aligned}
w_{i,t}&=\prod_{k=0}^{t}
\frac{\pi(a_{i,k}\mid h_{i,k})}{\mu(a_{i,k}\mid h_{i,k})},\\
\widehat V_{\mathrm{PDIS}}(\pi)
&=\frac1N\sum_{i=1}^{N}\sum_{t=0}^{T_i}
\gamma^t w_{i,t}r_{i,t}.
\end{aligned}
\label{eq:pdis}
\end{equation}
In Eq.~\eqref{eq:pdis}, \(N\) instead denotes the number of logged trajectories, \(T_i\) is the final time of trajectory \(i\), \(k\) indexes time inside the propensity product, \(t\) indexes the current reward, and \(r_{i,t}\) is the observed one-step reward. The quantity \(w_{i,t}\) is the cumulative importance weight through time \(t\), and \(\widehat V_{\mathrm{PDIS}}\) is the per-decision importance-sampling estimate. This \(w_{i,t}\) is an evaluation weight, distinct from the workload/interruptibility state component \(w_t\) in Eq.~\eqref{eq:state}; \(\gamma\) remains the discount factor defined in Eq.~\eqref{eq:cpomdp}.
Sequential doubly robust or model-based estimators can also control variance \cite{wang2017ope,liu2020ope,jiang2016dr}. These methods require consistent outcome definitions, support of the target policy under the behavior policy, no unrecorded sequential confounding conditional on logged history, and complete action, probability, and outcome logs; sensitivity analysis is needed when assumptions may fail. Importance weighting cannot identify actions outside the support of a deterministic log. Work on the intervention paradox further shows that a high failure-prediction AUROC can coexist with lower end-to-end success, because an intervention can rescue a failing trajectory or damage one that would have succeeded \cite{interventionparadox2026}.

\subsection{Risk, User Burden, and Reproducibility}

Risk evaluation should cover both \textbf{incidence} and \textbf{severity}: unauthorized-action rate, severity-weighted harm, near misses, successful undo or rollback, sensitive-data exposure, and explanation--action consistency. Let a larger selection score \(c\) indicate that intervention is more likely to be acceptable or beneficial, and let \(L\) be a prespecified loss on covered interventions, such as severe-risk loss \(L_R\) or a cost-weighted error. Selective-intervention studies should report, as threshold \(\tau\) varies,
\begin{equation}
\mathrm{Cov}(\tau)=\Pr(c\ge\tau),\qquad
\mathrm{Risk}(\tau)=\mathbb E[L\mid c\ge\tau],
\label{eq:coveragerisk}
\end{equation}
Here, \(\mathrm{Cov}(\tau)\) is the fraction of opportunities whose score reaches threshold \(\tau\), and \(\mathrm{Risk}(\tau)\) is the conditional mean loss among those covered interventions; probability and expectation are taken over the deployment-opportunity distribution. The scalar selection score \(c\) here is distinct from the cost vector \(\mathbf c\) in Eq.~\eqref{eq:cpomdp}.
rather than only the result at a best threshold. User burden should include interventions per unit time, question turns, rejection or ignoring, time to resume the primary task, and perceived interruption. A short-term click or ``like'' is not a proxy for long-term trust.

Reproducible experiments require fixed model versions, prompts, tool permissions, history windows, inference budgets, graders, and random seeds. LLM-based assessment should report the rubric, order randomization, agreement with human judgment, and sensitivity to alternative judge models. For real streams, authors should state whether retrieval leaks future information, whether memories are isolated between users, and whether privacy filtering changes the negative distribution. Public code and data links are one necessary condition, but do not replace a complete protocol or evidence from users.

\section{From Methods to Deployment Regimes}

\subsection{Three Regimes Share the Same Decision Variables}

Applications need not create another domain catalog. Table~\ref{tab:regimes} organizes work into three deployment regimes by observation continuity, time window, action reversibility, permission, and feedback delay. The threshold and admissible set of the same method change systematically across regimes.

\begin{table*}[t]
\centering
\caption{Mapping three deployment regimes to common decision variables. Dominant risks and permissions are constraints, not judgments of a domain's value.}
\label{tab:regimes}
\small
\setlength{\tabcolsep}{3.8pt}
\begin{tabularx}{\textwidth}{p{2.1cm}p{3.0cm}p{2.5cm}p{3.0cm}Y}
\toprule
Regime & State and timescale & Typical actions & Dominant constraints & Representative work and suitable design \\
\midrule
Digital workspace & Screens, documents, code, and cross-application events; seconds to days & Clarification, drafts, GUI/tool execution & Interleaved intents, account permissions, external side effects, rollback & PIRA, ProAgentBench, PersonalAlign, and Ambig-SWE \cite{pirabench2026,proagentbench2026,lyu2026personalalign,ambigswe2025}; reviewable \assist and tiered \act \\
Contextual and embodied environment & Egocentric video, wearables, and robot state; frames to minutes & Timely cues, navigation, collaborative actions & Perception delay, bystander privacy, physical safety, human workload & ESTP, ProMemAssist, Satori, and PACE \cite{eyeswideopen2025,promemassist2025,satori2025,pace2025}; risk-aware gating and action synchronization \\
High-stakes human service & Individual state in health, education, and psychological support; turns to months & Intervention messages, coaching questions, support policies, escalation & Professional scope, consent, long-term effects, vulnerable users, fairness & JITAI, SCALA, and counseling/support systems \cite{nahumshani2018jitai,jitaismoking2025,lastjitai2025,scala2026,cami2025,psyprobe2026}; conservative \assist, professional oversight, and longitudinal evaluation \\
\bottomrule
\end{tabularx}
\end{table*}

In digital workspaces, a richer \(\act\) set is appropriate when actions are sandboxed and tested, their impact can be previewed, and external side effects can be rolled back. Continuous screen capture nevertheless exposes information unrelated to the task, and reversibility is not costlessness. Software-engineering clarification shows that detecting underspecification, asking a useful question, and exploiting the answer are distinct bottlenecks \cite{ambigswe2025,clariti2026}. In contextual and embodied environments, timing and perception errors are coupled: late help may be useless, while an incorrect physical action may cause harm. Bayesian assistance and progress estimation incorporate uncertainty and human task progress into the gate \cite{bayesianproactive2024,pace2025}. HRI studies further distinguish anticipating a person's next step from taking initiative on that prediction and examine how proactive behavior affects teamwork \cite{roman2024,initiativeteams2024}. The proposed ``Goldilocks'' intervention window is a useful timing hypothesis, but presently a conceptual framework rather than evidence for a universal threshold \cite{goldilocks2025}. High-stakes human services often have delayed outcomes, so offline appropriateness scores cannot establish clinical, educational, or psychological benefit. Classroom studies and JITAI designs offer protocols closer to deployment but still require professional oversight and longitudinal controls \cite{scala2026,nahumshani2018jitai,jitaismoking2025}.

Emotional-support data and methods place affective trajectories, dialogue stage, and support strategy inside state--action construction: ESConv supplies strategy labels, while EmoDynamiX and ESCA model mixed-emotion/discourse dynamics and stage-aware strategy planning \cite{liu2021esconv,emodynamix2025,esca2026}. AFlow and LEKIA currently provide only preprint evidence on affective flow and a situated ``psychological world''; they are hypotheses for state construction, not evidence of psychological safety or clinical efficacy \cite{aflow2026,lekia2026}. In education, GenMentor, TRAVER, and Learning-to-Prompt respectively study goal guidance, turn-level verification, and adaptive prompting; as a companion paper, a Findings paper, and a preprint, they support mechanism or short-horizon comparisons rather than longitudinal learning effects \cite{genmentor2025,traver2025,learningtoprompt2026}.

\subsection{Encoding Safety in the Action Space and Interaction Design}

Equation~\eqref{eq:admissible} gives an abstract constraint, but systems also require operational safety mechanisms. \textbf{Tiered permission} separates reading, drafting, reversible modification, external communication, and irreversible commitment; authorization should bind an object, purpose, and duration. \textbf{Progressive autonomy} lets an agent degrade from \(\act\) to \(\assist\) or \(\ask\) under low confidence or high risk, rather than choose only between acting and not acting. \textbf{Recoverability} requires preview, impact scope, idempotent design, undo logs, and human takeover. \textbf{Auditable grounds} should faithfully record trigger evidence, material uncertainty, and permission checks instead of generating a post hoc account unrelated to the decision. \textbf{Data minimization} limits sensing, memory, and retrieval to information necessary and proportionate to a declared purpose and authorized service scope, together with provenance tracking and user access and deletion. Equation~\eqref{eq:advantage} is a utility model, not a legal or ethical necessity test.

MindGuard and WatchGuardian provide system examples of an on-device mental-health agent and user-defined just-in-time intervention, respectively, informing data minimization and personalized gating. WatchGuardian is currently a preprint, and neither study by itself establishes clinical benefit \cite{mindguard2025,watchguardian2025}.

These mechanisms alter the optimal policy rather than merely adding an ethical statement. Reversible action lowers \(C^{\mathrm{exec}}\), clear authorization enlarges \(\mathcal A_{\mathrm{perm}}(\bar\kappa_t)\), on-device processing lowers \(C^{\mathrm{priv}}\), and repeated confirmation raises \(C^{\mathrm{ask}}\). Safety design should therefore appear as a measurable variable in methods and evaluation, not as a principles list appended to the paper.

\subsection{Four Testable Questions Implied by Current Evidence}

\textbf{Q1: How can counterfactual intervention value be learned?} Current gates commonly learn a label for whether help is needed now, whereas Eq.~\eqref{eq:policyvalue} asks what intervention changes relative to silence. Where equipoise, ethics approval, and safety guardrails permit, a testable direction is to use micro-randomized trials, encouragement designs, or conservative exploration at real decision opportunities, log behavior probabilities, and measure user-endorsed net utility and long-term outcomes. Offline AUROC is an intermediate diagnostic rather than deployment evidence.

\textbf{Q2: How can need, content, and authorization be calibrated jointly?} Proposition~1 shows that changes in the conditional utility of candidate actions can induce distinct switching thresholds; it does not impose a fixed autonomy order on the four semantic modes. Future benchmarks can annotate the benefit--cost tradeoffs of multiple candidate actions under the same state, compare an independent need classifier with a fixed action against an approximate joint \(Q^*_{\boldsymbol\eta}(b,\bar\kappa,m,z,\ell)\) model, and report calibration, coverage--risk, and overreach. An improvement in content score without an improvement in net utility does not establish a better gate.

\textbf{Q3: How can an agent adapt safely under preference drift?} Accept and reject signals are selectively generated by the old policy, while long-term memory may retain obsolete goals. Evaluation should partition time and users to test drift detection, policy update, forgetting of stale preferences, and safety regression. At minimum it should jointly report post-adaptation benefit, old-scenario performance, authorization violations, and user deletion controls, rather than only memory question-answering accuracy.

\textbf{Q4: How can a common protocol cross regimes without erasing their constraints?} A single overall leaderboard hides differences in time windows, risk, and feedback. A more defensible design shares the decision-opportunity record, four-mode confusion matrix, event matching, and policy value, and separately reports interaction realism R, comparison design C, and human-outcome horizon H. Comparisons can then remain within digital, embodied, and high-stakes regimes. This preserves contextual constraints while testing whether the same formal claim holds across domains.

\section{Conclusion}

Proactive service is not the ability to generate an unsolicited sentence. It is a system's autonomous selection, under a genuine silence alternative and incomplete information, of a trigger, subgoal, question, or action not fully specified by the current directive. Formulating this choice as sequential decision making constrained by authorization and risk unifies timing, content, and delivery as a structured action. Waiting acquires option value; the pure information value of clarification depends on whether it changes later action value, while its total value may also include immediate interaction effects.

The representative studies reviewed here offer state estimation, policy planning, clarification learning, and streaming resources, but the evidence in Table~\ref{tab:benchmarks} is concentrated in constructed samples, offline replay, and short-term simulation. Higher trigger F1, more fluent advice, or longer memory is insufficient by itself to establish deployment benefit. A central next step is to learn the causal incremental value of intervention relative to silence, while constraining policies through calibrated uncertainty, explicit authorization, progressive autonomy, recoverable execution, and longitudinal user outcomes. Proactivity approaches trustworthy service when evaluation shows that the agent reliably waits under insufficient evidence, improves net utility when it intervenes, and leaves auditable grounds for the decision.

\balance
\bibliographystyle{IEEEtran}
\bibliography{refs}
\end{document}